\documentclass[final]{cvpr}

\usepackage{times}
\usepackage{epsfig}
\usepackage{graphicx}
\usepackage{amsmath}
\usepackage{amssymb}
\usepackage{caption}

\usepackage{array}

\newcommand{\rowfonttype}{}% Current row font
\newcommand{\rowfont}[1]{% Set current row font
   \gdef\rowfonttype{#1}#1%
}
\newcolumntype{L}[1]{>{\raggedright\let\newline\\\arraybackslash\hspace{0pt}}m{#1}}
\newcolumntype{C}[1]{>{\centering\let\newline\\\arraybackslash\hspace{0pt}}m{#1}}
\newcolumntype{R}[1]{>{\raggedleft\let\newline\\\arraybackslash\hspace{0pt}}m{#1}}
\usepackage{multirow}
\newcommand{\colwidthA}{3.0cm}
\newcommand{\colwidthB}{1.3cm}
\newcommand{\colwidthC}{1.4cm}
\usepackage[pagebackref=true,breaklinks=true,colorlinks,bookmarks=false]{hyperref}

\def\cvprPaperID{} % *** Enter the CVPR Paper ID here
\def\confYear{CVPR 2021}

\begin{document}

%%%%%%%%% TITLE
\title{Meticulous Object Segmentation}

\author{Chenglin Yang\textsuperscript{1}\footnotemark, Yilin Wang\textsuperscript{2}, Jianming Zhang\textsuperscript{2}, He Zhang\textsuperscript{2}, Zhe Lin\textsuperscript{2}, Alan Yuille\textsuperscript{1}\\
\textsuperscript{1}Johns Hopkins University\quad\textsuperscript{2}Adobe Inc.\\
{\tt\small \{chenglin.yangw,alan.l.yuille\}@gmail.com \quad \{yilwang,jianmzha,hezhan,zlin\}@adobe.com}}

\twocolumn[{
\renewcommand\twocolumn[1][]{#1}
\maketitle
\newcommand{\colwidthF}{0.125\textwidth}
\begin{center}
 \centering
 \small
 \setlength{\tabcolsep}{0.0pt}
 \begin{tabular}{C{\colwidthF}C{\colwidthF}C{\colwidthF}C{\colwidthF}C{\colwidthF}C{\colwidthF}C{\colwidthF}C{\colwidthF}}
     \multicolumn{8}{c}{\includegraphics[width=1.0\textwidth]{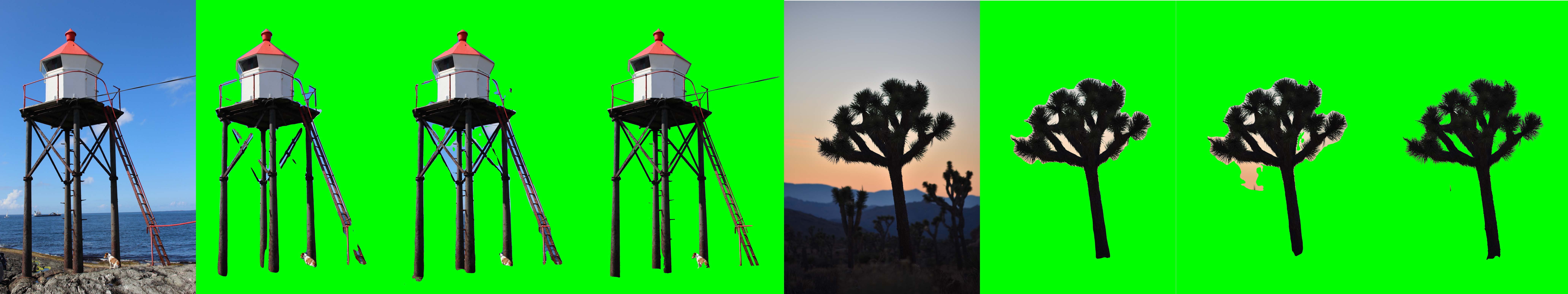}}
     \\
     Input &
     EGNet \cite{zhao2019egnet} &
     MINet \cite{pang2020multi} &
     Ours &
     Input &
     EGNet \cite{zhao2019egnet} &
     MINet \cite{pang2020multi} &
     Ours 
 \end{tabular}
 \captionof{figure}{A visual comparison of different methods by compositing the foreground object on green background. See more examples in Figure~\ref{fig: visual comparisons_8}, \ref{fig: visual comparisons_9}, \ref{fig: visual comparisons_10} and \ref{fig: visual comparisons_11}. Best viewed with zoom-in.
 }
 \label{fig:Teaser}
\end{center}
}]

\maketitle

\renewcommand{\thefootnote}{\fnsymbol{footnote}}
\setcounter{footnote}{1} 
\footnotetext{Work done while an intern at Adobe.}
\setcounter{footnote}{1} 
\renewcommand*{\thefootnote}{\arabic{footnote}}

%%%%%%%%% ABSTRACT
\begin{abstract}
Compared with common image segmentation tasks targeted at low-resolution images, higher resolution detailed image segmentation receives much less attention. 
In this paper, we propose and study a task named Meticulous Object Segmentation (MOS), which is focused on segmenting well-defined foreground objects with elaborate shapes in high-resolution images (\eg 2k - 4k). To this end, we propose the MeticulousNet which leverages a dedicated decoder to capture the object boundary details. Specifically, we design a Hierarchical Point-wise Refining (HierPR) block to better delineate object boundaries,
and reformulate the decoding process as a recursive coarse to fine refinement of the object mask.
To evaluate segmentation quality near object boundaries, we propose the Meticulosity Quality (MQ) score 
considering both the mask coverage and boundary precision. 
In addition, we collect a MOS benchmark dataset including 600 high quality images with complex objects. We provide comprehensive empirical evidence showing that MeticulousNet can reveal pixel-accurate segmentation boundaries and is  superior to state-of-the-art methods for high-resolution object segmentation tasks. 
The code is publicly available\footnote{\href{https://github.com/Chenglin-Yang/MOS_Meticulous-Object-Segmentation}{\textcolor{blue}{https://github.com/Chenglin-Yang/MOS\_Meticulous-Object-Segmentation}}}.

\end{abstract}

\section{Introduction}

With mobile camera devices and digital imaging tools being used more and more extensively, current applications, such as image editing and manipulation, become more popular and bring forward large demands on techniques of foreground object segmentation ~\cite{borji2019salient,shen2016automatic}. Importantly, these real-world use cases require the techniques to run on high resolution images.

As the resolution of an image going up, more details are revealed by the increasing number of pixels, such as animal furs, human hairs, insect antennas, flower stamens, cavities inside jewelry, handrails at boats, etc. These phenomena around object boundaries increase the difficulty for foreground segmentation at high resolution to pursue good qualities. Unfortunately, most of existing object segmentation works \cite{zhao2019egnet,pang2020multi} focus on low resolution images, where fine-grained details constituting the object boundaries are not well represented and ignored in the model design. On the other hand, directly processing high resolution image such as 4K resolution will be memory expensive and hard to recover those details \cite{wang2020deep,cheng2020cascadepsp}.  To address these challenges and investigate model capabilities in the computer vision community, we propose and study a task named Meticulous Object Segmentation (MOS).

To distinguish MOS from the traditional task Salient Object Detection (SOD), let's start with some examples shown in Figure~\ref{fig:Teaser}. The watchtower along the coast is supported by a architecture with cross-linked wooden poles which together with the ladder and handrails make the object boundaries very complicated. The two state of the art SOD models~\cite{zhao2019egnet,pang2020multi} are able to segment the object body but fail to capture the boundaries. It is the same scenario for the pine tree. The thick branches criss-cross, making the models unable to discriminate the foreground and backgrounds near the branches.  These observations indicate that existing SOD methods may not sufficient for MOS. Inherited from visual saliency \cite{itti1998model,borji2012state} in computer vision, SOD aims at detecting salient object in an image, while how to precisely capture object details is not fully considered. By contrast, MOS is designated for meticulously segmenting out complex foreground objects. For example, as shown in the Figure ~\ref{fig:Teaser}, our method for MOS successfully deal with these boundary and structure details.

\textbf{Task Datasets:} 
To ensure fair comparisons and enhance the importance of model developments, we limit the training data to be a collection of current public SOD datasets which is illustrated in Section~\ref{sec: training datasets}. To evaluate MOS methods, we release a testing set MOS600 consisting of 600 high resolution images, whose ground truth masks are carefully annotated. These images contain objects with complicated shapes, and we quantitatively measure their complexity in Section~\ref{sec: M1000}.

\textbf{Task Metrics:} the evaluation metric for MOS is supposed to consider both the object bodies and boundaries of segmentation masks. 
Traditional metrics IoU and mBA can evaluate the performance in these two perspectives separately. Besides, 
we invent Meticulosity Quality (MQ), depicted in Section~\ref{sec: MQ}, 
to measure the boundary quality in a moderate way by also considering the mask coverage,
as required by MOS. All these metrics are important to measure MOS performance.

\begin{figure}[t]
    \centering
    \includegraphics[width=0.46\textwidth]{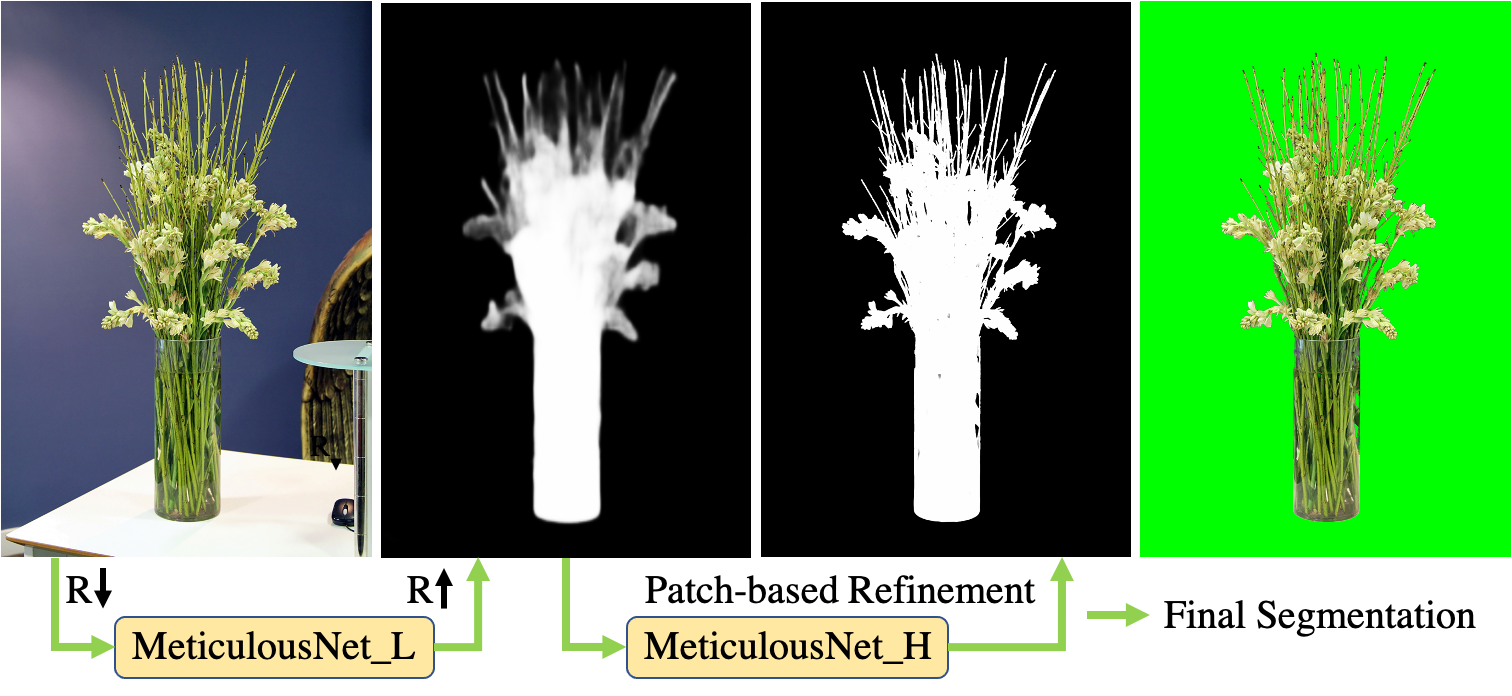}
    \caption{Our framework pipeline. The arrows indicate the resolution (R) change.
    The input image is first resized to a low resolution. MeticulousNet\_L predicts a coarse foreground score map which is then resized back. MeticulousNet\_H refines it to a high quality result which is then binarzied as the segmentation mask. The final image shows the composite of the segmented foreground and green background. }
    \label{fig: framework}
\end{figure}

To tackle this challenging task, we propose \textbf{MeticulousNet}, standing for the network equipped with our designed decoder recovering detailed segmentation boundaries. It includes the following designs: First, it is internally embedded with hierarchical point-wise refining blocks in the decoder layers. This structure enables the network to self-evaluate the uncertainty maps and take them as guidance for the local refinements. Second, we integrate a recursive mask-wise global refinements into the decoder coupled with local modules, boosting the performance by iteratively improving the quality of object masks in higher resolution. As shown in Figure ~\ref{fig: framework}, our framework consists of a low resolution segmentation model and a high resolution refinement model. In spite of the different functionalities of these models, they share the same architecture that we designed, MeticulousNet. We use the \_L and \_H to discriminate them, indicating low and high quality of the model inputs.

We experimentally study MOS in Section~\ref{sec: MeticulousNet_L} and~\ref{sec: MeticulousNet_H}, where we evaluate MOS methods on HRSOD~\cite{zeng2019towards} and MOS600 using both traditional metrics and MQ. Our contributions can be summarized as following: 

(1) We propose Meticulous Object Segmentation (MOS), a foreground segmentation task on high resolution images with complex object boundaries. To evaluate MOS performance, we release MOS600 and invent Meticulosity Quality representing the natural challenges and requirements of integrating body and boundary segmentation in the high resolution scenarios. (2) We propose MeticulousNet with a powerful decoder incorporating the hierarchical local refinements and recursive global refinements under the internal unsupervised spatial guidance. This decoder is developed to recover complex foreground boundaries which is used to tackle MOS. 

\section{Related Work}

Meticulous Object Segmentation (MOS) is proposed as a challenging task. Its difficulty lies in the combination of foreground segmentation and detailed boundary segmentation in the scenario of processing objects with complex shapes at high resolution. In this section, we review traditional tasks and models that tackle these problems separately. 

\subsection{Salient Object Detection}

Salient Object Detection (SOD) is to distinguish the most visually salient objects from the backgrounds in images. Early solutions involve the usage of hand-crafted features~\cite{itti1998model,klein2011center,wei2012geodesic,zhang2013saliency,yang2013saliency,zhu2014saliency,cheng2014global,zhang2016ranking,zhang2017saliency,borji2019salient}. The advent of deep convolutional neural networks (CNN) and their feature representations with enormous flexibility~\cite{krizhevsky2012imagenet, simonyan2014very, he2016deep, szegedy2015going, huang2017densely} takes SOD methods into CNN-based stage~\cite{zhao2015saliency,li2015visual,wang2015deep}, which is followed by the works~\cite{wang2016saliency,liu2016dhsnet,luo2017non,zhang2017amulet,zhang2018bi,liu2018picanet,zhang2018progressive} based on FCN and U-Net~\cite{long2015fully,ronneberger2015u}. Recently, Zhao \textit{et al}. proposed EGNet designed with a salient edge detection module which is aimed at assisting in object detection~\cite{zhao2019egnet}. Later in the work of Pang \textit{et al}., they designed MINet equipped with aggregate-interaction and self-interaction modules to better handle scale variations~\cite{pang2020multi}.

In spite of good performances of these state of the art SOD models, they are only suitable for segmenting the foregrounds out of low resolution images. Because at this scale, the boundary details are missing due to the insufficient number of pixels, and are therefore not significant enough to affect the segmentation qualities. In the experimental comparison between a three-branch SOD framework~\cite{zeng2019towards} for larger input size with our MOS method (Section~\ref{sec: MeticulousNet_H}), we found a large performance drop, $>11\%$ in boundary accuracy at HRSOD in which the objects have more than $9$ times less boundary complexities than MOS600
(Section~\ref{sec: M1000}). 
Both the goal and model developments of SOD focus on object body segmentation, which prevents SOD techniques from being a powerful tool dealing with high resolution images and complex object boundaries.

\subsection{Segmentation Refinement}

In order to improve the quality of image segmentation results, refinement techniques have been proposed and explored. Traditional methods include the integration of graphical models~\cite{krahenbuhl2011efficient,chen2014semantic,zheng2015conditional,lin2016efficient} with deep neural networks. Later on, individual refining modules appeared~\cite{peng2017large,zhang2019canet}. One of the design directions is creating generic plug-in modules enabling internal iterative refinements inside the models. RefineNet, proposed by Lin \textit{et al}, consists of a cascade of RefineNet units~\cite{lin2017refinenet}. These units connecting with difference levels of features perform multi-path refinements in the upsampling process inside the network. Kirillov \textit{et al}. proposed PointRend module that iteratively selects the most uncertain regions and refines these regions increasing the resolution to a higher value
according to point-wise fine and coarse features~\cite{kirillov2020pointrend}. Another direction is creating refiners capable of refining masks that predicted by architecture-agnostic models. CascadePSP proposed by Cheng \textit{et al}. adopts a global-local patch-based recursive refining pipeline and achieves the state of the art refiner performance~\cite{cheng2020cascadepsp}.

The majority of the quality improvements brought about by this independent refiner are on the segmentation mask boundaries, especially for high resolution images. Its cascade network structure makes it possible that better performances are achieved by increasing inference times. Specifically, the input of CascadePSP is the concatenation of the raw image and three previously predicted masks that can be either the same or different in a coarse-to-fine fashion depending on the iteration index. However, the dependence of the cascade structure on the existence of previously predictions limits its usage on the refining task. In this work, we break this limitation, and design a decoder (Section~\ref{sec: methods}) that can enjoy cascade benefits in both the refining and direct prediction processes.

\section{MeticulousNet}
\label{sec: methods}

Our MOS framework consists two parts with same architecture design:  a low resolution  MeticulousNet\_L segmentation model and a high resolution MeticulousNet\_H refinement model.
As shown in Figure ~\ref{fig: framework}, We first feed the low resolution image to MeticulousNet\_L to obtain low resolution mask, then Meticulous\_H will crop the high resolution image patch along with low resolution mask to recover mask details in the original 
resolution. Since both MeticulousNet\_L and MeticulousNet\_H share the same network design, we describe MeticulousNet in the following.

\subsection{Overview}
\label{sec: architecture overview}

\begin{figure}[t]
    \centering
    \includegraphics[width=0.9\linewidth]{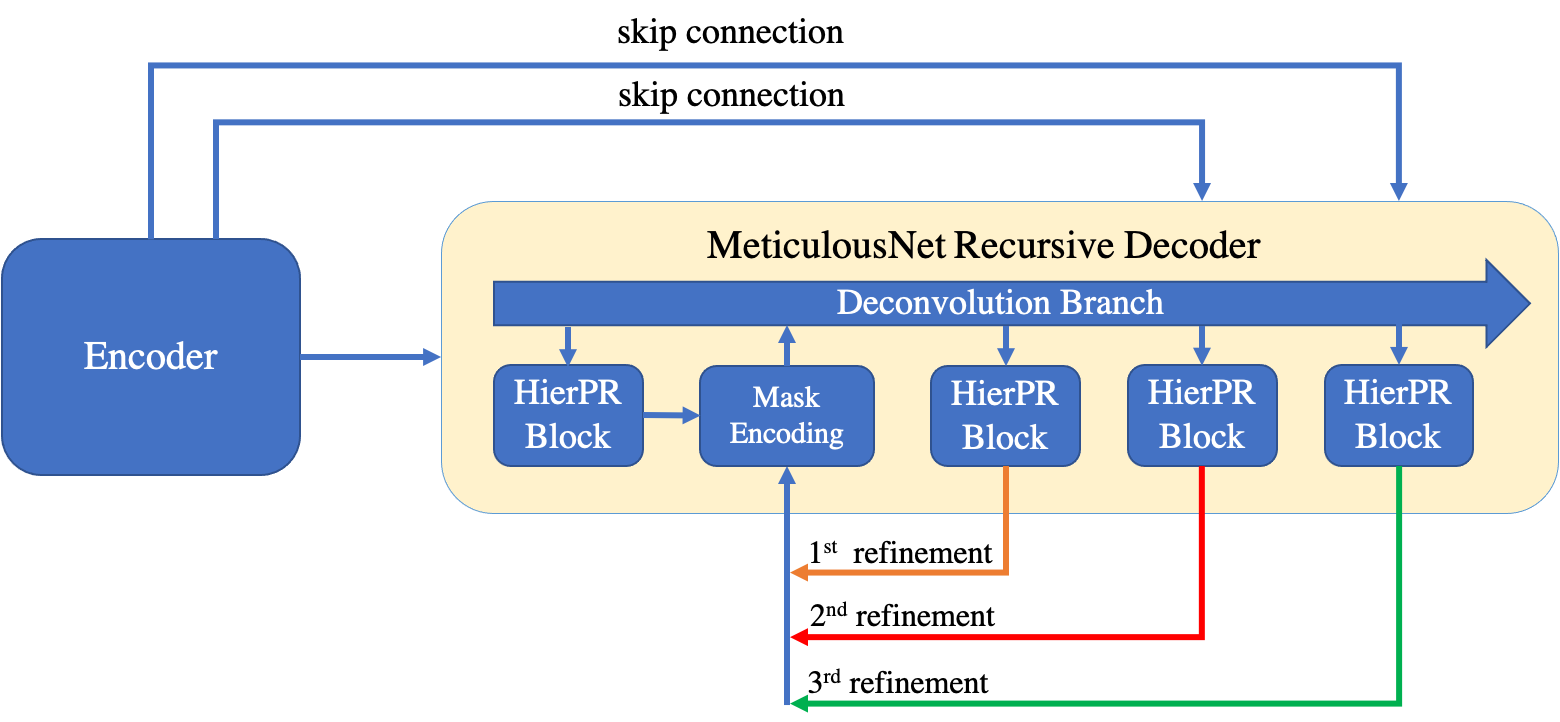}
    \caption{MeticulousNet overview. Orange, red and green arrows represent the recursive processes of the decoder.}
    \label{fig: MeticulousNet overview}
\end{figure}

The overview of MeticulousNet is shown in Figure~\ref{fig: MeticulousNet overview}. It adopts encoder-decoder structure. The encoder can be the feature extraction modules from other networks, in this paper we use the encoder of PSPNet~\cite{zhao2017pyramid} with ResNet50~\cite{he2016deep} and MobileNetV3~\cite{howard2019searching} backbones. What differentiates MeticulousNet from other models lies in the decoding process. We invent a recurrent decoder achieving local and global self-refinements.

\subsection{HierPR Block}

\begin{figure}[t]
    \centering
    \includegraphics[width=0.9\linewidth]{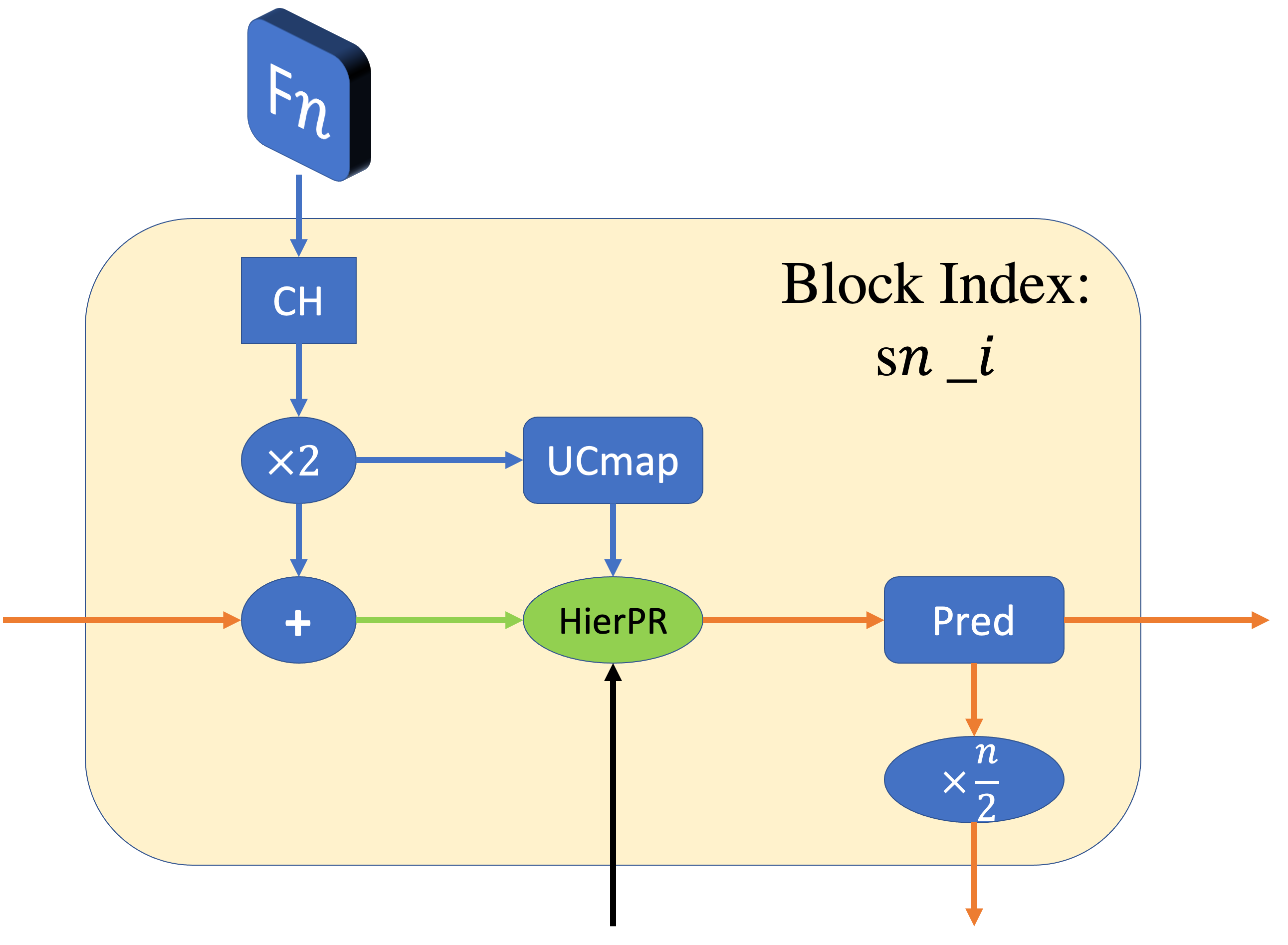}
    \caption{HierPR block. F$n$ represents the feature maps with output stride $n$ w.r.t. the input size. CH represents the coarse head predicting result, which is formed by two $1 \times 1$ convolutional layers. 
    UCmap means the uncertainty map which is used as guidance to perform refinements. HierPR is a three-layer perceptron and Pred is the prediction map after HierPR refinements. HierPR block needs three inputs: pre-specified feature maps from the encoder, feature maps from the deconvolutional layers, predictions from previous HierPR block, which are incicated by the black, blue and orange arrows. $i$ is the cycle index in decoder shown in Figure~\ref{fig: recursive_decoder}. 
    }
    \label{fig: HierPR block}
\end{figure}

Fine object boundaries do not cover large portion of the image area, which drives the need of selecting regions containing these details with correct spatial positions and limited number of pixels in order to perform effective local refinements. Based on~\cite{kirillov2020pointrend}, we design a Hierarchical Point-wise Refining (HierPR) block. It performs local refinements at selected pixels in the low-level feature maps with pixel inputs from high level feature maps in a hierarchical way.

The Basic PointRend (PR) module \cite{kirillov2020pointrend} is a three-layer perceptron. 
At each point to perform refinements, the input is the concatenation of the feature vector from pre-specified feature activations and the prediction vector from coarse prediction map at that spatial location.
The potential resolution inconsistency is dealt with by interpolation. The coarse prediction is also fed into the intermediate layers for better performance. Since it is computationally infeasible to apply PR at all possible points, an estimated uncertainty map is introduced to provide guidance and limit computation costs. The values in this map are calculated as the absolute distances between the coarse prediction and $0.5$ in this binary setting. Then a sorting is performed and a pre-configured number of most uncertain locations are selected to apply refinements. The initial reason why we apply PointRend is to utilize its structure constrain to
make the model learn to locate where the object boundaries are in an unsupervised way. 

Based on this point-wise processing unit, we propose a hierarchical refinement structure within the basic element HierPR block, which is visualized in Figure~\ref{fig: HierPR block} and~\ref{fig: recursive_decoder}. From a macro view, our decoder is an unification of two branches. The first one recovers the resolution of feature maps through deconvolutional layers to the strides of $8$, $8$, $4$ and $2$ gradually. The second one refines the coarse outputs coming from the deconvolution branch, or an averaged combination of outputs from both branches (stride $4$ and $2$) , to the strides of $4$, $4$, $2$ and $1$ respectively, by means of point-wise refinements. 
From a micro view, the skip connection pattern differs in these two branches. The last two deconvolutional layers are connected to the first block and first convolutional layer in the encoder, using features from high to low levels. While in the refining branch, all the HierPR modules are only connected to the first block, using the same features to refine all the predictions. These features are relatively low-level when refining the stride $8$ prediction but relatively high-level when refining the stride $2$ prediction. This contrastive design on the skip connections between two branches enables the HierPR block to provide detail information at low resolution and semantic guidance at high resolution adaptively when doing refinements.

\subsection{Recursive Decoder}
\label{sec: recursive decoder}

\begin{figure}[t]
    \centering
    \includegraphics[width=0.46\textwidth]{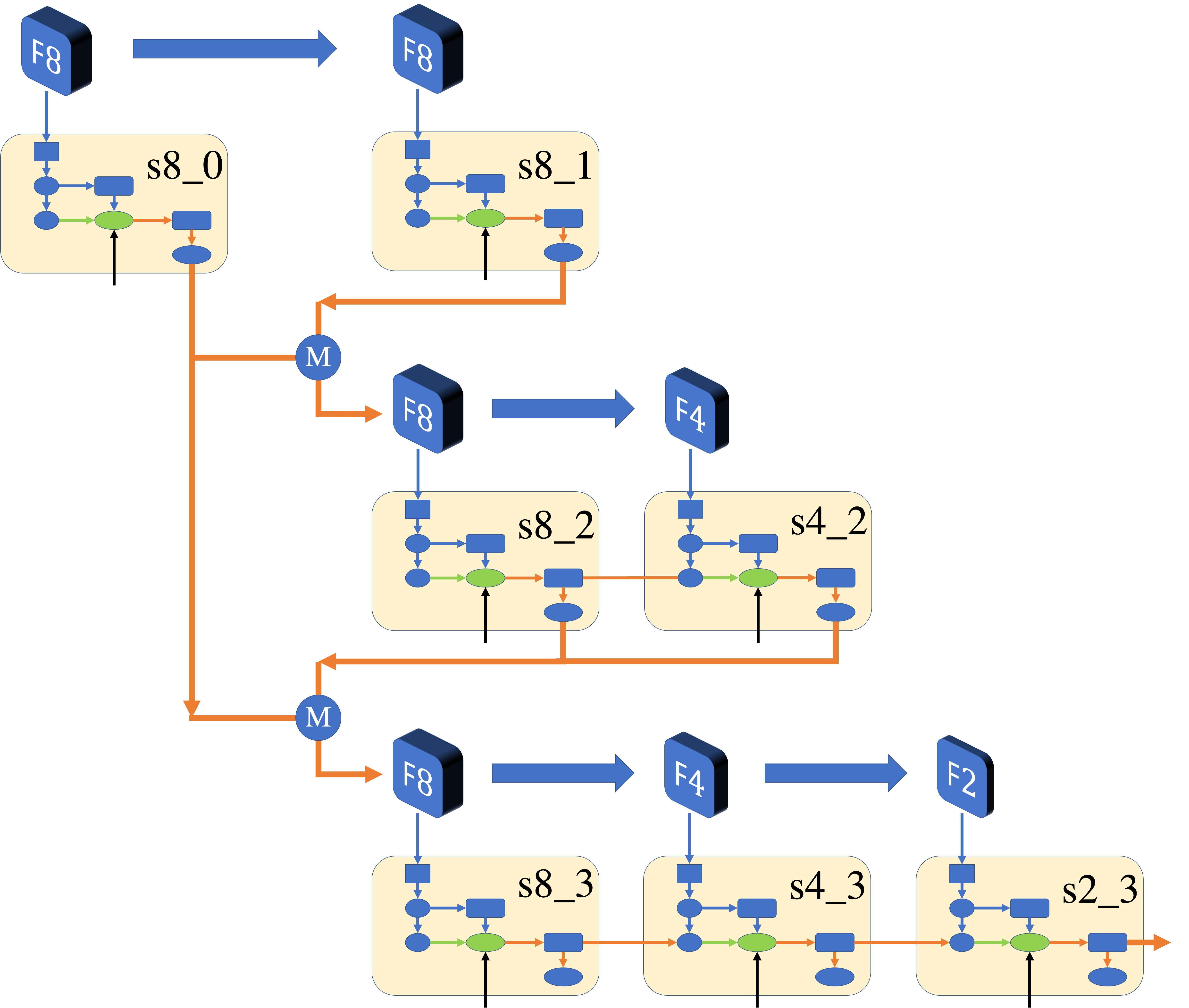}
    \caption{Recursive decoder. The upsampling process proceeds from left to right. The top branch is deconvolution branch and the bottom is refinement branch with HierPR blocks. s$n$\_$i$ represents the block index where $n$ and $i$ represents the output stride of the input feature maps from the top branch to the block and the internal cycle index where this block belongs to. M is the mask encoding layer.
    The details of each block is shown in Figure~\ref{fig: HierPR block}. The orange arrows show the recursive process. Modules in green share weights globally and those in blue share weights when vertically aligned.
    See Section~\ref{sec: recursive decoder} for details}
    \label{fig: recursive_decoder}
\end{figure}

Complementary to hierarchical local refinements, we design our global refinements in a recursive manner. Inspired by~\cite{cheng2020cascadepsp} where a complete network is proposed to refine a pre-computed segmentation mask iteratively, we enable our decoder to internally perform global refinements mask-wisely.

As shown in Figure~\ref{fig: recursive_decoder}, the global refining pipeline in our decoder can be abstracted into two parts. First, it initializes a stride $4$ mask by applying an one-time HierPR on a stride $8$ coarse features to stabilize the training. Second, there exists a three-cycle upsampling process which is represented in three rows, respectively.
These cycles are of different lengths, recovering the resolution to the stride $4$, $2$ and $1$ sequentially via cascade of HierPR blocks. 
The temporary predicted masks at the end of each internal cycle are concatenated in sequence with the initial stride $4$ prediction, which is out of these cycles, as the input to the mask encoding layer. This encodes the coarse-to-fine variations on the predicted masks, which is then utilized to guide global refinements in the next cycle. 
During training, the supervision are added onto all the intermediate and final predictions, with a weighted combination of binary cross entropy, L1 and L2 losses. (See section~\ref{sec: implementation details})

\newcommand{\colwidthF}{0.1\textwidth}
\begin{table}[t]
 \centering
 \small
 \setlength{\tabcolsep}{0.0pt}
 \begin{tabular}{C{\colwidthF}C{\colwidthF}C{\colwidthF}C{\colwidthF}C{\colwidthF}}
     \multicolumn{5}{c}{\includegraphics[width=0.5\textwidth]{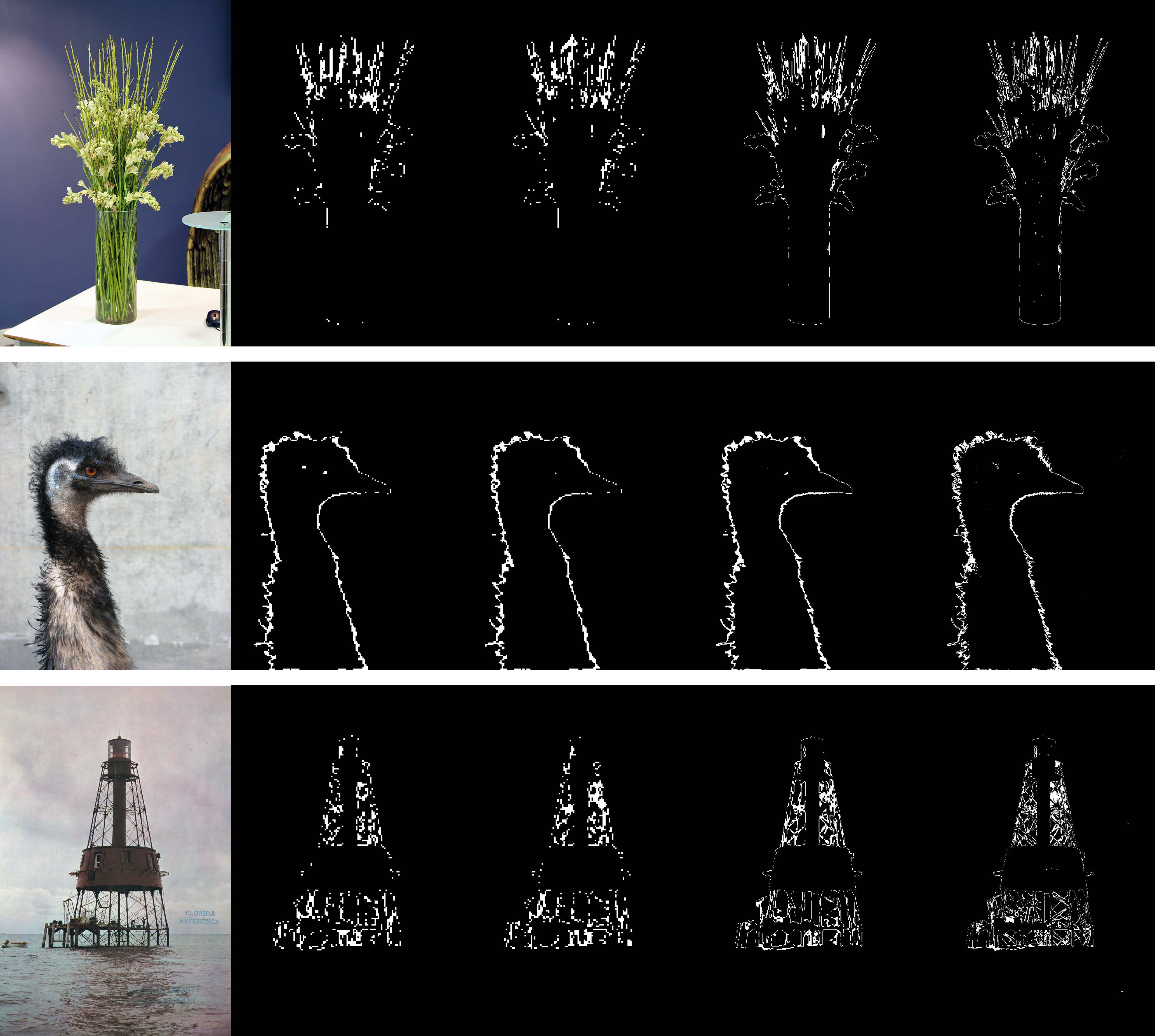}}
     \\
     \rowfont{\scriptsize}input &\rowfont{\scriptsize}s8\_0 &\rowfont{\scriptsize} s8\_3 &\rowfont{\scriptsize} s4\_3 &\rowfont{\scriptsize} s2\_3
     \\
     \multicolumn{5}{c}{(a) Uncertainty maps learnt progressively and without supervision.}
     \\
     \multicolumn{5}{c}{\includegraphics[width=0.5\textwidth]{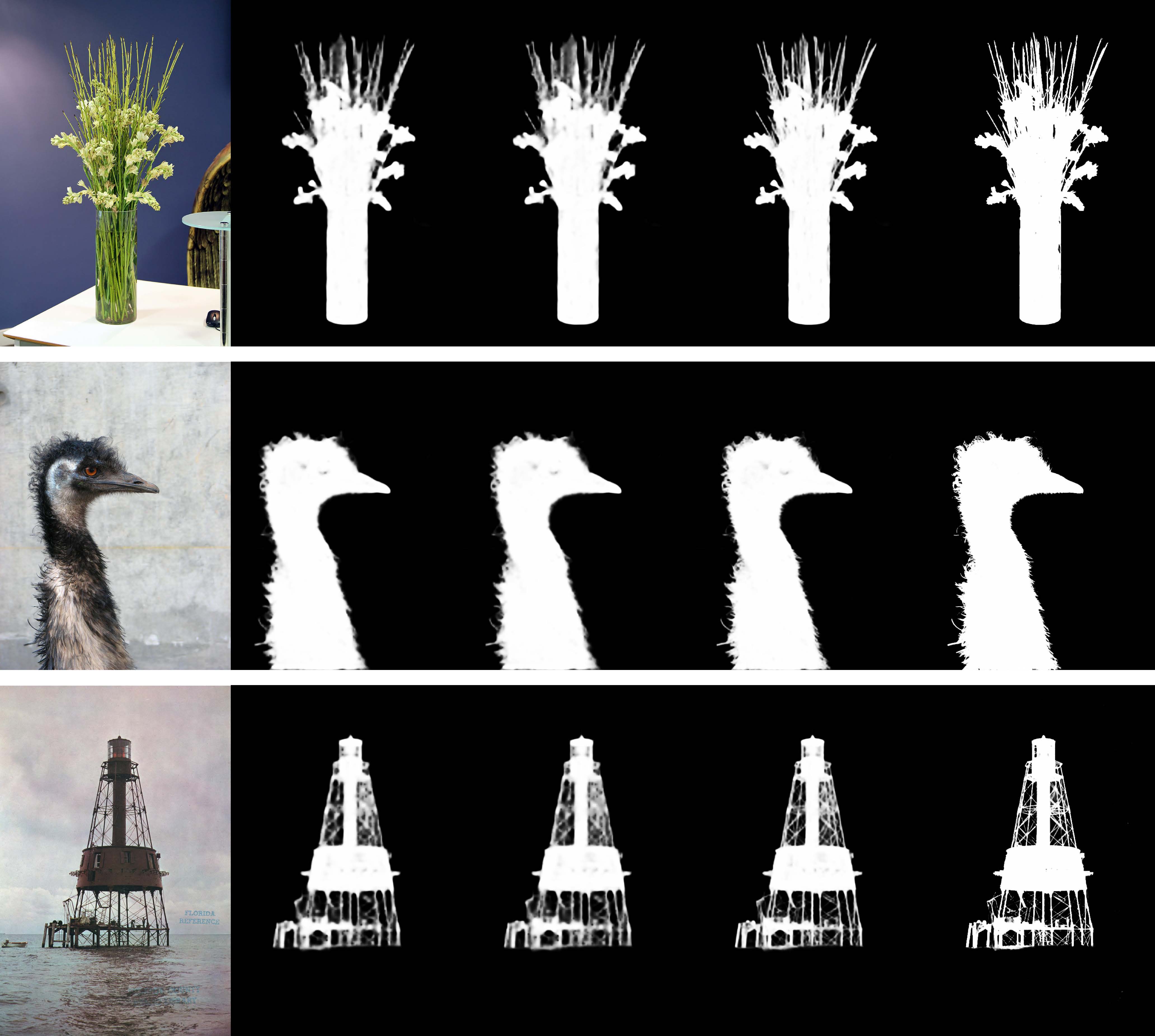}}
     \\
     \rowfont{\scriptsize}input &\rowfont{\scriptsize}s8\_0 &\rowfont{\scriptsize} s8\_1 &\rowfont{\scriptsize} s4\_2 &\rowfont{\scriptsize} s2\_3
     \\
     \multicolumn{5}{c}{(b) Intermediate progressive predictions.}
     \\
 \end{tabular}
 \captionof{figure}{Internal Visualizations of MeticulousNet\_H. The first and second subfigures represent the generated uncertainty maps (UCmap) and prediction maps (Pred) in HierPR blocks. The index s$n$\_$i$ corresponds to Figure~\ref{fig: recursive_decoder}. Best viewed with zoom-in.}
 \label{fig: internal visualization}
\end{table}

Combining both the local and global processes, there are $7$ pairs of predictions and $7$ uncertainty maps generated by $7$ HierPR blocks. 
We use S$n$\_$i$ to index these blocks where $n$ and $i$ represents the output stride of input feature maps coming from the deconvolution branch and index of internal cycles where this block is. $i$ ranges from $0$ to $3$ with $0$ represents the first prediction outside the $3$ latter cycles.

Figure~\ref{fig: internal visualization} visually records the updates on the uncertainty maps and predictions through these HierPR blocks when the model is fed with an unseen example. The uncertainty maps are grasping the boundaries more and more accurately meaning that our network not only learns a spatial guidance of where to apply refinements correctly but also continuously improves this guidance quality during the upsampling process, which is done in an unsupervised way. Also as shown in the second subfigure, the prediction masks at the end of the recursive cycles provide masks with increasing quality and confidence both locally and globally. All of these behaviors correspond well to our design intention. 

\section{Experiments}
\label{sec: experiments}

In this section, we propose a metric to quantitatively evaluate Meticulous Object Segmentation (MOS) and release a dataset named MOS600. It contains objects with complex boundaries that are quantitatively measured and serves as the benchmark testing set of MOS. Besides, we also evaluate our methods using traditional metrics and dataset HRSOD~\cite{zeng2019towards}. We compare our methods with the corresponding state of the art models in both the low resolution segmentation and high resolution refinement tasks.

\subsection{Meticulosity Quality}
\label{sec: MQ}

\renewcommand{\colwidthF}{0.200\textwidth}
\begin{table*}[t]
 \centering
 \small
 \setlength{\tabcolsep}{0.0pt}
 \begin{tabular}{C{\colwidthF}C{\colwidthF}C{\colwidthF}C{\colwidthF}C{\colwidthF}}
     \multicolumn{5}{c}{\includegraphics[width=1.0\textwidth]{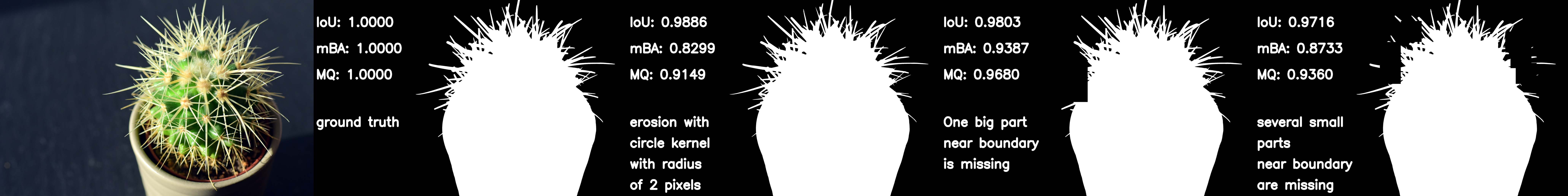}}
     \\
     (a) & 
     (b) & 
     (c) & 
     (d) & 
     (e) 
 \end{tabular}
 \captionof{figure}{A visual illustration of MQ with IoU, mBA. (a), (b) are the original image and its ground truth mask (GT). (c), (d) and (e) are obtained by perturbing GT. Best viewed with zoom-in. 
 }
 \label{fig: MQ illustration}
\end{table*}

The boundary quality of segmentation masks can be measured by mean Boundary Accuracy (mBA) proposed in~\cite{cheng2020cascadepsp}. For a given input image with size $w$ and $h$, $N$ radii are sampled in $[1, \frac{(w+h)}{300}]$ with uniform intervals, which are $r_{1},...,r_{N}$ in ascending order. For each $r_{i}$, a boundary region $b_{i}$ is calculated as the difference between the dilation and erosion of the ground truth mask with a circle kernel whose diameter is $2r_{i}+1$. Then mBA is calculated as the average pixel accuracy in these $N$ areas. 
However, to evaluate MOS performance, the segmentation quality of the object body should also be considered. Therefore we invent Meticulosity Quality (MQ), which is calculated as:
\begin{align*}
& MQ = \frac{1}{2}P_{b_{N}}^{\text{O}} + \frac{1}{2N} \sum_{i=1}^{N}{P_{b_{i}}^{\text{I}}}
\end{align*}
where $P_{b_{i}}^{\text{I}}$ and $P_{b_{i}}^{\text{O}}$ is the pixel accuracy inside and outside the region $b_{i}$.
The second term comes from mBA, and we set $N=5$ as well. 
The reason why mBA is calculated as the average accuracy on $5$ different small boundary areas is to ensure the robustness of the measurement for boundaries, which is followed by us. The regions 
outside $b_{i}$ can be considered to contain object bodies. 
Since the area outside is far larger than these small boundary areas, and thus the measurement outside is not affected by several pixel shifts and do not have stability issues. So we just take area outside $b_{5}$ to measure body segmentation. 
MQ of a perfect mask will be $1$. 

All the metrics IoU, mBA and MQ are important to evaluate Meticulous Object Segmentation (MOS). We provide illustrations in Figure~\ref{fig: MQ illustration}. 
In subfigure (c) where the mask are eroded by $2$ pixels, mBA is decreased by a large margin but the visual quality of the segmentation is not affected that severely. 
In subfigure (d), one big part near object boundary is missing and the visual quality is more harmed than small erosion, but IoU is still as high as that in subfigure (c). Same scenario happens in subfigure (e) where several small parts near boundary are missing, IoU can not well represent missing boundary details while mBA shows large performance degradation although main object body is segmented correctly. 
MQ provide a perspective to reflect the quality of 
boundary segmentation in a moderate way. 
Since IoU and mBA are dedicated in different perspectives, all these metrics are necessary to measure performances of MOS methods.

\subsection{Meticulous Benchmark Testing Dataset}
\label{sec: M1000}

We release a dataset of $600$ high resolution images with complex boundaries, MOS600. To measure the boundary complexity, we exploit $C_{IPQ}$~\cite{osserman1978isoperimetric} as used in~\cite{zeng2019towards}. Before the calculation, we crop the patch including the foreground and resize it to be a square as a calibration process.
We compare it with the testing set of HRSOD, consisting of $400$ carefully annotated high resolution images. As shown in Table~\ref{tab: boundary complexity}, the object boundaries in MOS600 are 
more than $9$
times more complex, which shows that MOS600 is a suitable benchmark test set for MOS.

\renewcommand{\colwidthA}{3.0cm}
\renewcommand{\colwidthB}{2.7cm}
\begin{table}[!btp]
\centering{
\setlength{\tabcolsep}{0.16cm}
\begin{tabular}{|C{\colwidthA}|C{\colwidthB}|}
\hline
Dataset & $C_{IPQ}$ ($ \times 10^{-3}$) $\downarrow$\\
\hline\hline
HRSOD   & $11.34$ \\ 
\hline
MOS600 & $\;\; 1.24$ \\ 
\hline
\end{tabular}}
\caption{Boundary complexity. Lower $C_{IPQ}$ indicates higher complexity. MOS600 is more than $9$ times more complex than HRSOD.}
\label{tab: boundary complexity}
\vspace{-0.0cm}
\end{table}

\subsection{Training Datasets}
\label{sec: training datasets}

We train our model using existing datasets. For the low resolution model MeticulousNet\_L, we use the training set of DUTS~\cite{wang2017learning} containing $10553$ images which have complicated scenes with objects in various scales and locations. For the high resolution model MeticulousNet\_H, we exploit the combination of MSRA-10K~\cite{cheng2014global}, DUT-OMRON~\cite{yang2013saliency}, ECSSD~\cite{shi2015hierarchical}, and FSS-1000~\cite{li2020fss} with $36572$ images in total, which is used by~\cite{cheng2020cascadepsp}. All of these collections contribute to a class-agnostic dataset with objects of diverse semantic categories and various annotation qualities. To enhance the supervision on object details, we add the binarized DIM~\cite{xu2017deep} to the collections. We merge the training and testing sets of DIM and remove the smoke categories, resulting in $463$ unique foregrounds. During training when a DIM foreground is sampled, we first compose it with a background image randomly selected from MS COCO~\cite{lin2014microsoft} and then randomly binarize its alpha map using a threshold randomly generated in the range of $0.1$ - $0.5$ to get the ground truth segmentation mask. For each training epoch, there are $46300$ such compositions.

\subsection{Implemenation Details}
\label{sec: implementation details}

At low resolution, MeticulousNet only uses Binary Cross Entropy loss (BCE) applied on the intermediate and final outputs. The learning rate is adjusted using a cosine annealing schedule, starts from $2.5\times 10^{-3}$ and decreases to $0$ in $60000$ iterations. The input image is resized to $336$, and batch size is $20$. Data augmentations involve random horizontal flips, rotations, and color jittering as ued in~\cite{pang2020multi}. At high resolution, we adopt patch-based training used in ~\cite{cheng2020cascadepsp}. A combination of BCE, L1 and L2 losses are utilized with a triple weights ($w_{b}$, $w_{1}$, $w_{2}$). 
Corresponding to the $7$ HierPR blocks in Table~\ref{fig: recursive_decoder}, we use $(1.00, 1.00, 1.00)$ for S$8\_0$, S$8\_1$, S$8\_2$, S$8\_3$; $(0.50, 0.25, 0.25)$ for S$4\_2$, S$4\_3$; $(0.00, 1.00, 1.00)$ for S$2\_3$. Gradient loss on the final prediction with a weight of $5$ is also added. The learning rate starting with $2.25 \times 10^{-4}$ is step-wisely decayed by a factor $10$ after iterations $22500$ and $37500$ in a total of $60000$. We use a batch size of $24$. The inputs are concatenations of $224\times 224$ patches cropped from the original images and their corresponding ground truth masks that are randomly perturbed to have an IoU in the range $0.8$ - $1.0$ with the unperturbed ones. At both resolutions, we set HierPR to refine $10\%$ pixels in the feature maps.
Before training at high resolution, the weights are initialized from the best training epoch on HRSOD at low resolution.

\subsection{MeticulousNet at Low Resolution}
\label{sec: MeticulousNet_L}

The goal of MOS is to segment objects from high resolution images, and the design of MeticulousNet aims at capturing complex boundaries. However, since our framework outputs a coarse mask and then refines it, we perform experiments to investigate how well MeticulousNet performs at low resolution. In this case, the resolution challenge disappears and boundary details are not significant for the foreground segmentation. We name our model Meticulous\_L and compare it with two state of art models for Salient Object Detection (SOD), EGNet~\cite{zhao2019egnet} and MINet~\cite{pang2020multi}. We use the model checkpoints released by the authors for fair comparison.
For testing sets, we resize HRSOD and MOS600 to the size $336 \times 336$. The predictions on these images by the models are used as the initial masks in the refinement stage which is studied in Section~\ref{sec: MeticulousNet_H}. Since all comparing methods use DUTS-TR as training set, we also report the comparisons on DUTS-TE.

The experimental results are reported in Table~\ref{Tab:low_res_DTUS-TE}, \ref{Tab:low_res_HRSOD} and ~\ref{Tab:low_res_MOS600}. On DUTS-TE, Meticulous\_L achieves the highest MQ, IoU and comparable mBA with EGNet. MINet has a slightly lower value in MAE but gets worst score otherwise. On HRSOD, Meticulous\_L shows a small degrade $0.09\%$ in mBA compared with EGNet, but gives the best performance on all other metrics. 
EGNet is able to handle foreground boundaries properly, and this can attribute to the fact that EGNet has a module predicting object boundary edges and thus receive extra supervision from object edges in the training process. However, when the object boundary complexities increase, its advantage 
disappear. On MOS600, MeticulousNet\_L provides the best performance on all the metrics. MINet does not perform well in mBA on all datasets with an average of more than $1.3\%$ performance gap with others. Together with the performances on DUTS-TE and HRSOD, MeticulousNet\_L consistently provides the best 
performance on
object body and boundary segmentation, which is reflected by the highest MQ, highest IoU, and competitive mBA. These properties at low resolution hint the large potential of MeticulousNet to process high quality data.

\renewcommand{\colwidthA}{2.3cm}
\renewcommand{\colwidthB}{0.95cm}
\begin{table}[!htp]
\centering{
\small
\setlength{\tabcolsep}{0.08cm}
\begin{tabular}{|C{\colwidthA}|C{\colwidthB}|C{\colwidthB}|C{\colwidthB}|C{\colwidthB}|C{\colwidthB}|C{\colwidthB}|C{\colwidthB}|C{\colwidthB}| C{\colwidthB}|}
\hline
Method & MAE $\downarrow$
& S-m $\uparrow$ & IoU $\uparrow$ & mBA $\uparrow$ & MQ $\uparrow$ \\
\hline\hline
EGNet & $3.91$ 
& $\mathbf{88.70}$ & $78.22$ & $\mathbf{66.44}$ & $81.79$ \\
\hline
MINet & $\mathbf{3.69}$ 
& $88.45$ & $78.14$ & $64.76$ & $80.98$ \\
\hline
MeticulousNet\_L & $3.76$ 
& $88.60$ & $\mathbf{78.40}$ & $66.42$ & $\mathbf{81.84}$ \\
\hline
\end{tabular}}
\caption{Low resolution foreground segmentation on DUTS-TE. (Units: $1\times 10^{-2}$)}
\label{Tab:low_res_DTUS-TE}
\end{table}

\renewcommand{\colwidthA}{2.3cm}
\renewcommand{\colwidthB}{0.95cm}
\begin{table}[!htp]
\centering{
\small
\setlength{\tabcolsep}{0.08cm}
\begin{tabular}{|C{\colwidthA}|C{\colwidthB}|C{\colwidthB}|C{\colwidthB}|C{\colwidthB}|C{\colwidthB}|C{\colwidthB}|C{\colwidthB}|C{\colwidthB}| C{\colwidthB}|}
\hline
Method & MAE $\downarrow$
& S-m $\uparrow$ & IoU $\uparrow$ & mBA $\uparrow$ & MQ $\uparrow$ \\
\hline\hline
EGNet & $4.05$ 
& $89.47$ & $80.19$ & $\mathbf{66.18}$ & $81.51$ \\
\hline
MINet & $3.58$
& $89.59$ & $80.49$ & $64.82$ & $80.96$ \\
\hline
MeticulousNet\_L & $\mathbf{3.46}$ 
& $\mathbf{89.75}$ & $\mathbf{80.59}$ & $66.09$ & $\mathbf{81.74}$ \\
\hline
\end{tabular}}
\caption{Low resolution foreground segmentation on HRSOD. (Units: $1\times 10^{-2}$)}
\label{Tab:low_res_HRSOD}
\end{table}

\renewcommand{\colwidthA}{2.3cm}
\renewcommand{\colwidthB}{0.95cm}
\begin{table}[!htp]
\centering{
\small
\setlength{\tabcolsep}{0.08cm}
\begin{tabular}{|C{\colwidthA}|C{\colwidthB}|C{\colwidthB}|C{\colwidthB}|C{\colwidthB}|C{\colwidthB}|C{\colwidthB}|C{\colwidthB}|C{\colwidthB}| C{\colwidthB}|}
\hline
Method & MAE $\downarrow$
& S-m $\uparrow$ & IoU $\uparrow$ & mBA $\uparrow$ & MQ $\uparrow$ \\
\hline\hline
EGNet & $6.89$ 
& $82.21$ & $70.50$ & $60.54$ & $77.83$ \\
\hline
MINet & $6.55$ 
& $81.94$ & $69.94$ & $59.30$ & $77.27$ \\
\hline
MeticulousNet\_L & $\mathbf{6.40}$ 
& $\mathbf{83.38}$ & $\mathbf{72.18}$ & $\mathbf{60.72}$ & $\mathbf{78.16}$ \\
\hline
\end{tabular}}
\caption{Low resolution foreground segmentation on MOS600. (Units: $1\times 10^{-2}$)}
\label{Tab:low_res_MOS600}
\end{table}

\subsection{MeticulousNet for MOS}
\label{sec: MeticulousNet_H}

\renewcommand{\colwidthA}{3.5cm}
\renewcommand{\colwidthB}{1.1cm}
\begin{table*}[!htp]
\centering{
\small
\setlength{\tabcolsep}{0.08cm}
\begin{tabular}{|C{\colwidthA}||C{\colwidthB}|C{\colwidthB}|C{\colwidthB}|C{\colwidthB}|C{\colwidthB}||C{\colwidthB}|C{\colwidthB}|C{\colwidthB}|C{\colwidthB}|C{\colwidthB}|C{\colwidthB}|C{\colwidthB}|C{\colwidthB}|C{\colwidthB}|C{\colwidthB}|C{\colwidthB}|C{\colwidthB}| C{\colwidthB}|}
\hline
Dataset & \multicolumn{5}{c||}{HRSOD} & \multicolumn{5}{c|}{MOS600} \\
\hline
Method & MAE $\downarrow$
& S-m $\uparrow$ & IoU $\uparrow$ & mBA $\uparrow$ & MQ $\uparrow$ & MAE $\downarrow$
& S-m $\uparrow$ & IoU $\uparrow$ & mBA $\uparrow$ & MQ $\uparrow$ \\
\hline\hline
EGNet & $4.06$ 
& $89.39$ & $80.10$ & $64.76$ & $80.91$ & $7.04$ 
& $81.87$ & $69.99$ & $61.25$ & $78.76$ \\
\hline
MINet & $3.61$ 
& $89.57$ & $80.41$ & $63.85$ & $80.59$ & $6.70$ 
& $81.69$ & $69.59$ & $60.13$ & $78.33$ \\
\hline
GSN+APS+LRN+GLFN & $2.98$ 
& $89.67$ & $80.40$ & $62.26$ & $80.11$ & ---- 
& ---- & ---- & ---- & ---- \\
\hline
EGNet + cascadePSP & $3.56$ 
& $89.34$ & $81.70$ & $72.32$ & $84.66$ & $5.88$ 
& $82.59$ & $73.12$ & $67.67$ & $82.04$ \\
\hline
MINet + cascadePSP & $2.98$ 
& $89.67$ & $82.64$ & $72.32$ & $84.82$ & $5.70$ 
& $83.00$ & $73.47$ & $67.25$ & $81.99$ \\
\hline
EGNet + cascadePSP\textbf{*} & $3.51$ 
& $89.29$ & $81.66$ & $72.41$ & $84.72$ & $5.56$ 
& $83.58$ & $74.69$ & $70.27$ & $83.40$ \\
\hline
MINet + cascadePSP\textbf{*} & $3.23$ 
& $89.98$ & $\mathbf{82.76}$ & $72.56$ & $84.94$ & $5.34$ 
& $84.13$ & $75.19$ & $69.96$ & $83.40$ \\
\hline
MeticulousNet\_(L+H) & $\mathbf{2.97}$ 
& $\mathbf{90.00}$ & $82.56$ & $\mathbf{73.64}$ & $\mathbf{85.59}$ & $\mathbf{5.08}$ 
& $\mathbf{85.08}$ & $\mathbf{76.43}$ & $\mathbf{72.09}$ & $\mathbf{84.46}$ \\
\hline
\end{tabular}}
\caption{Meticulous object segmentation in high resolution, Note that boundary complexity in MOS is more than $9$ times higher than HRSOD. (Units: $1\times 10^{-2}$)}
\label{Tab:high_res_HRSOD}
\end{table*}

In this section, we show the experimental results on MOS. Based on predictions from Meticulous\_L, which is described in Section~\ref{sec: MeticulousNet_L}, our framework consists of another MeticulousNet with the same architecture as high resolution refiner, Meticulous\_H. To ensure fair comparisons, we set our strong baselines as combinations of the state of the art (SOTA) SOD model and SOTA refiner. We use EGNet and MINet to predict coarse masks, and cascadePSP~\cite{cheng2020cascadepsp} for refinements. We use the best released checkpoints. Since MeticulousNet\_H includes binarized DIM in the training set as mentioned in Section~\ref{sec: training datasets}, we also re-train a model,  cascadePSP\textbf{*}, using the same training data. All the other training and testing procedures are ensured the same.
We perform comparisons on HRSOD testing set and MOS600 the images of which are never seen by the models. Another baseline methods are GLF networks proposed with HRSOD, and we use their released predicted masks for comparisons.

The experimental results are shown in Table~\ref{Tab:high_res_HRSOD}. On HRSOD, 
GLF networks show a large performance gap with 
our proposed strong baselines. 
Comparing the baseline methods with cascadePSP and cascadePSP\textbf{*}, we do not observe clear benefits brought by the addition of binarized DIM data. For example, based on the predictions by MINet, cascadePSP\textbf{*} increases both IoU and mBA by $0.12\%$ and $0.24\%$, respectively, compared with raw cascadePSP. However, on top of predictions by EGNet, cascadePSP\textbf{*} increases mBA by $0.08\%$ but decreases IoU by $0.04\%$. Comparing MeticulousNet with all the other baselines, it shows a small degrade in IoU, $0.20\%$ lower compared with cascadePSP\textbf{*} on top of MINet, but has a more than $1\%$ increase in mBA compared with all the baselines. 
It also achieves the highest performance on all the other metrics.

On MOS600, since the object boundary complexities are increased by more than $9$ times than HRSOD as shown in Section~\ref{sec: M1000}, the differences among different methods are enlarged. MeticulousNet provides the best performance on all the metrics among all the methods. Comparing it with the off-the-shelf cascadePSP on top of SOD models, it increase IoU, mBA and MQ by at least $2.96\%$, $4.42\%$ and $2.42\%$, respectively. Comparing the baseline methods, we observe a benefit from the addition of the binarized DIM data. On top of both EGNet and MINet, cascadePSP\textbf{*} improves the performance in MQ by around $1.4\%$ compared with cascadePSP. Comparing this strongest baseline with MeticulousNet, there are consistent improvements on all the metrics. It improves IoU, mBA and MQ by $1.24\%$, $2.13\%$ and $1.06\%$. All these experiments verify the superior capability of MeticulousNet to deal with the segmentation of both foreground body and boundary, as required by MOS. See the visual comparisons in Figure \ref{fig: visual comparisons_8}, \ref{fig: visual comparisons_9}, \ref{fig: visual comparisons_10} and \ref{fig: visual comparisons_11}.

\renewcommand{\colwidthA}{2.5cm}
\renewcommand{\colwidthB}{1.1cm}
\begin{table}[t]
\centering{
\setlength{\tabcolsep}{0.16cm}
\begin{tabular}{|C{\colwidthA}|C{\colwidthB}|C{\colwidthB}|C{\colwidthB}|}
\hline
Method & IoU $\uparrow$ & mBA $\uparrow$ & MQ $\uparrow$ \\
\hline\hline
cascadePSP & $92.22$ & $74.67$ & $86.74$ \\
\hline
MeticulousNet & $\mathbf{92.23}$ & $74.85$ & $86.84$ \\
\hline
cascadePSP\textbf{*} & $92.02$ & $74.34$ & $86.56$ \\
\hline
MeticulousNet\_H & $91.96$ & $\mathbf{75.32}$ & $\mathbf{87.03}$ \\
\hline
\end{tabular}}
\caption{Refinement performance on BIG dataset.All the networks refine the same prediction masks predicted by DeeplabV3+~\cite{chen2018encoder}. Both the training of cascadePSP\textbf{*} and MeticulousNet\_H include the usage of the binarized DIM data.}
\label{Tab: refinement exp}
\end{table}

To better demonstrate the effectiveness of MeticulousNet\_H, we compare it with cascasdePSP directly in the refinement task on high resolution dataset BIG~\cite{cheng2020cascadepsp} studied in their paper. We download the segmentation masks pre-computed by the authors using semantic segmentation model DeepLabV3+~\cite{chen2018encoder}. They are used as inputs for all the refinement models. The results are shown in Table~\ref{Tab: refinement exp}. Comparing cascadePSP and MeticulousNet, it is observed they show comparable performance in IoU but MeticulousNet shows better performance in boundary segmentation by an increase of $0.18\%$ in mBA, which is also reflected in MQ.
After adding the binarized DIM data, we observe a performance degrades for both cascadePSP\textbf{*} and MeticulousNet\_H in IoU, which is about $0.20\%$ decrease. This means this additional training data harm the object body segmentation. For mBA, the scenarios are different for these two networks. MeticulousNet\_H can better utilize this high quality data to improve its segmentation results around object boundaries, and by contrast, cascadePSP\textbf{*} behave worse.

\renewcommand{\colwidthF}{0.15\textwidth}
\begin{table*}[t]
 \centering
 \small
 \setlength{\tabcolsep}{0.0pt}
 \begin{tabular}{C{\colwidthF}C{\colwidthF}C{\colwidthF}C{\colwidthF}C{\colwidthF}C{\colwidthF}}
     \multicolumn{6}{c}{\includegraphics[width=0.90\textwidth]{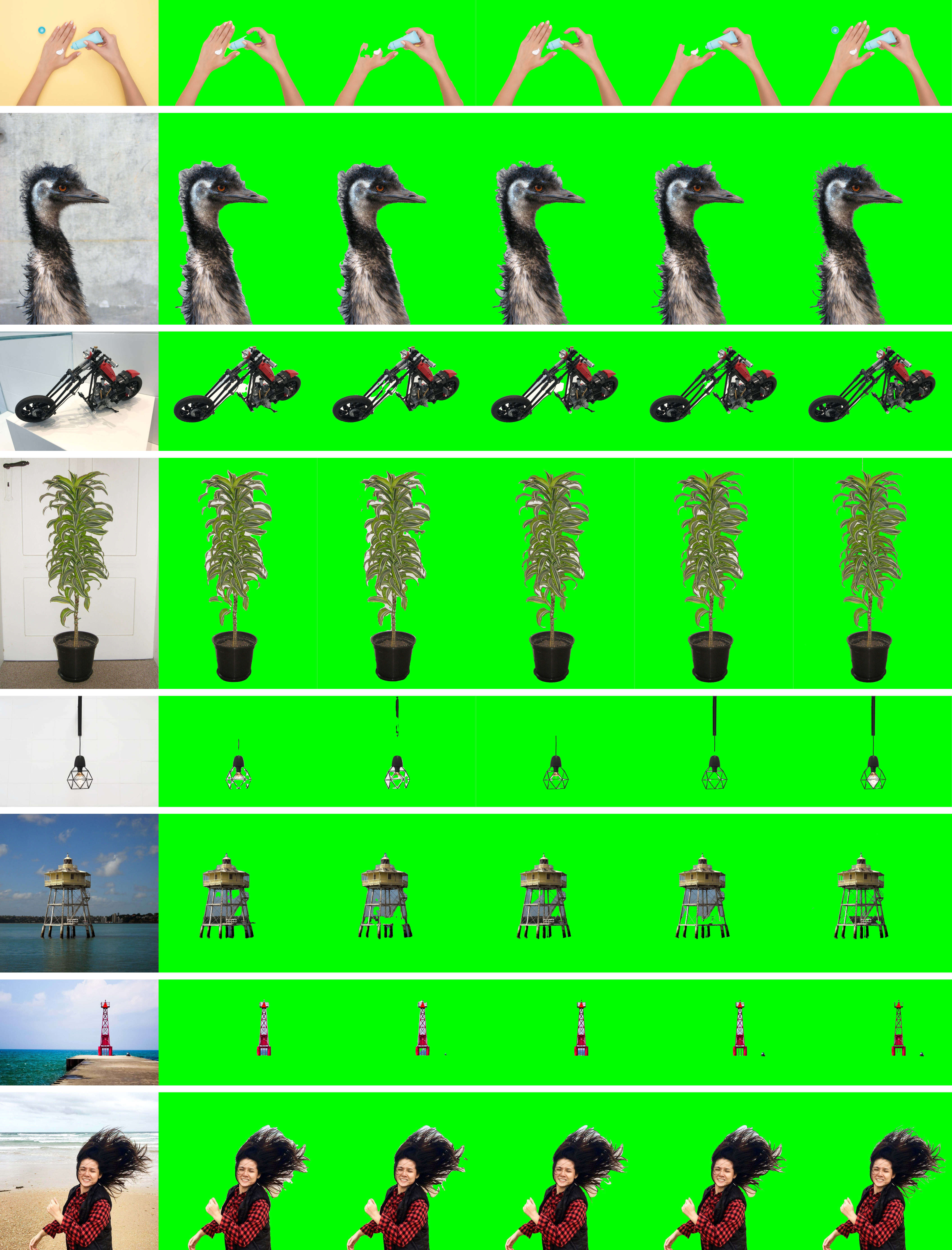}}
     \\
     \rowfont{\scriptsize}Input &\rowfont{\scriptsize} EGNet &\rowfont{\scriptsize} MINet &\rowfont{\scriptsize} EGNet+cascadePSP &\rowfont{\scriptsize} MINet+cascadePSP &\rowfont{\scriptsize} MeticulousNet\_L+H
 \end{tabular}
 \captionof{figure}{Visual comparisons among baselines and MeticulousNet (1).}
 \label{fig: visual comparisons_8}
\end{table*}

\renewcommand{\colwidthF}{0.15\textwidth}
\begin{table*}
 \centering
 \small
 \setlength{\tabcolsep}{0.0pt}
 \begin{tabular}{C{\colwidthF}C{\colwidthF}C{\colwidthF}C{\colwidthF}C{\colwidthF}C{\colwidthF}}
     \multicolumn{6}{c}{\includegraphics[width=0.90\textwidth]{figure_pool/figure_pool_5_pure_figure_opt_1200_Page_3.jpg}}
     \\
     \rowfont{\scriptsize}Input &\rowfont{\scriptsize} EGNet &\rowfont{\scriptsize} MINet &\rowfont{\scriptsize} EGNet+cascadePSP &\rowfont{\scriptsize} MINet+cascadePSP &\rowfont{\scriptsize} MeticulousNet\_L+H
 \end{tabular}
 \captionof{figure}{Visual comparisons among baselines and MeticulousNet (2).}
 \label{fig: visual comparisons_9}
 \vspace{1.0cm}
\end{table*}

\renewcommand{\colwidthF}{0.15\textwidth}
\begin{table*}
 \centering
 \small
 \setlength{\tabcolsep}{0.0pt}
 \begin{tabular}{C{\colwidthF}C{\colwidthF}C{\colwidthF}C{\colwidthF}C{\colwidthF}C{\colwidthF}}
     \multicolumn{6}{c}{\includegraphics[width=0.90\textwidth]{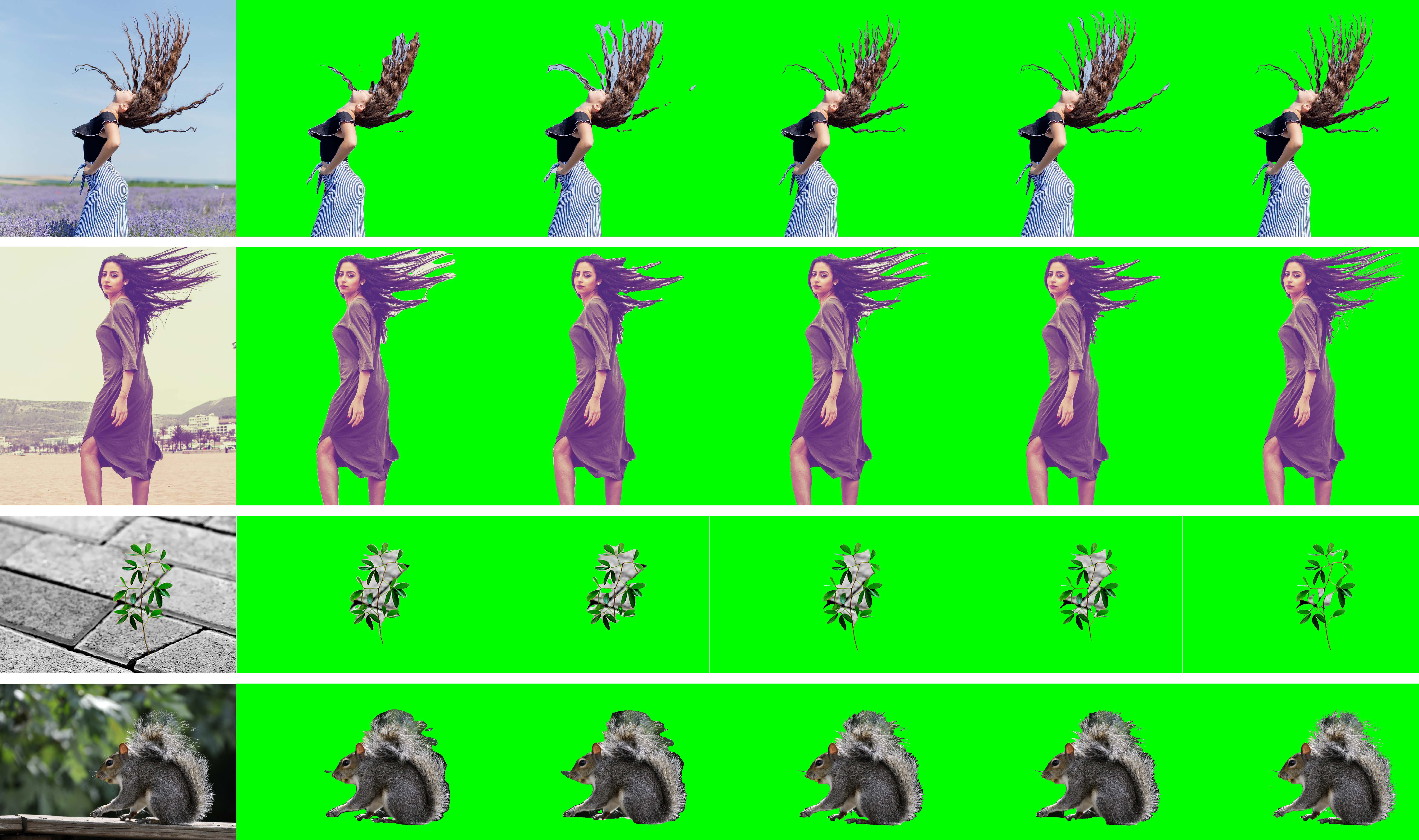}}
     \\
     \rowfont{\scriptsize}Input &\rowfont{\scriptsize} EGNet &\rowfont{\scriptsize} MINet &\rowfont{\scriptsize} EGNet+cascadePSP &\rowfont{\scriptsize} MINet+cascadePSP &\rowfont{\scriptsize} MeticulousNet\_L+H
 \end{tabular}
 \captionof{figure}{Visual comparisons among baselines and MeticulousNet (3).}
 \label{fig: visual comparisons_10}
\end{table*}

\renewcommand{\colwidthF}{0.125\textwidth}
\begin{table*}[!htp]
 \centering
 \small
 \setlength{\tabcolsep}{0.0pt}
 \begin{tabular}{C{\colwidthF}C{\colwidthF}C{\colwidthF}C{\colwidthF}C{\colwidthF}C{\colwidthF}C{\colwidthF}C{\colwidthF}}
     
     \multicolumn{8}{c}{\includegraphics[width=1.00\textwidth]{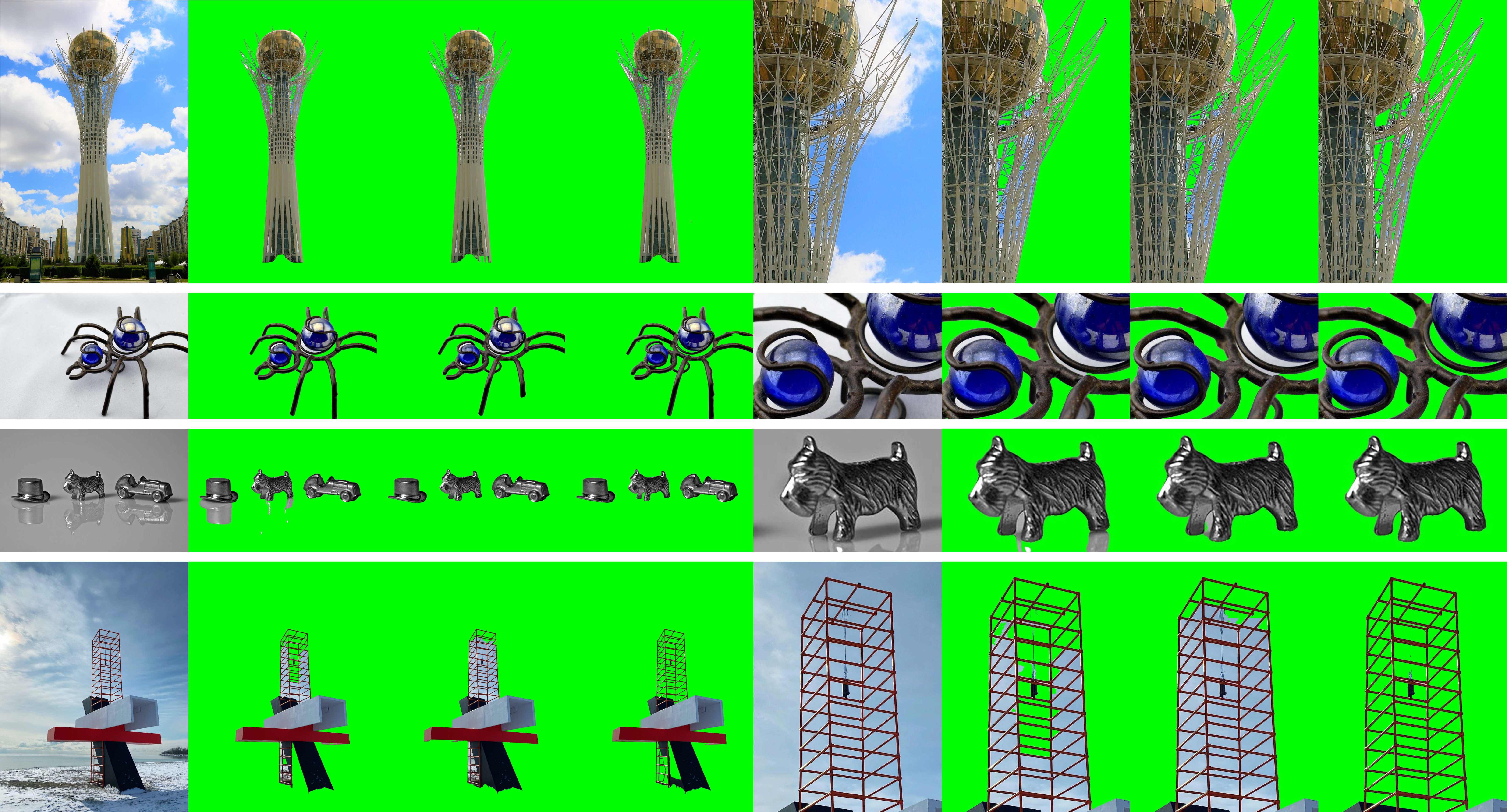}}
     \\
     \rowfont{\scriptsize}Input &\rowfont{\scriptsize} EGNet+cascadePSP\textbf{*} &\rowfont{\scriptsize} MINet+cascadePSP\textbf{*} &\rowfont{\scriptsize} MeticulousNet\_L+H & \rowfont{\scriptsize}Input &\rowfont{\scriptsize} EGNet+cascadePSP\textbf{*} &\rowfont{\scriptsize} MINet+cascadePSP\textbf{*} &\rowfont{\scriptsize} MeticulousNet\_L+H
 \end{tabular}
 \captionof{figure}{Visual comparisons among baselines and MeticulousNet (4). The last $4$ columns are the cropped patches from the first $4$ columns.}
 \label{fig: visual comparisons_11}
\end{table*}

\renewcommand{\colwidthF}{0.15\textwidth}
\begin{table*}
 \centering
 \small
 \setlength{\tabcolsep}{0.0pt}
 \begin{tabular}{C{\colwidthF}C{\colwidthF}C{\colwidthF}C{\colwidthF}C{\colwidthF}C{\colwidthF}}
 
     \multicolumn{6}{c}{\includegraphics[width=0.90\textwidth]{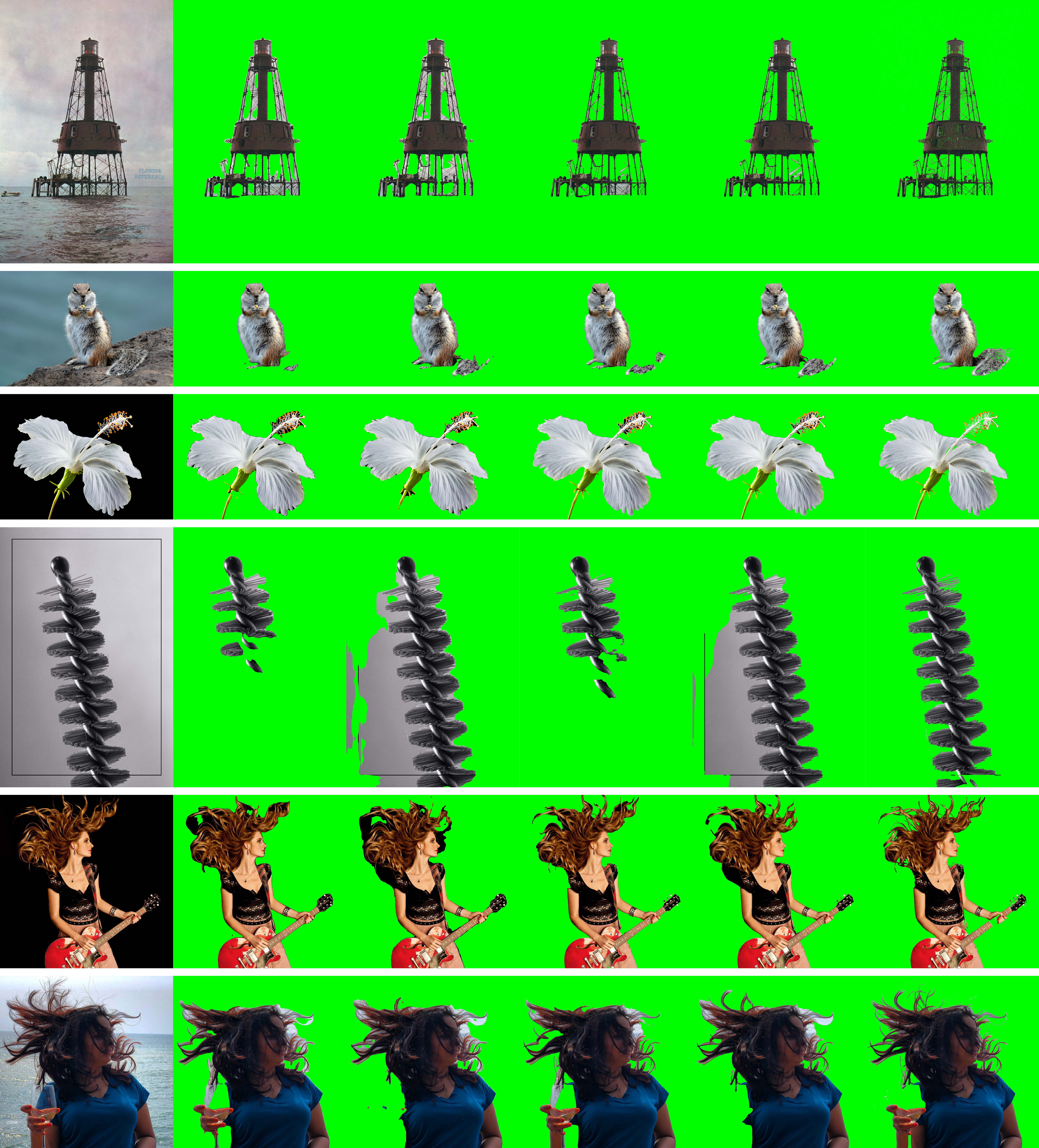}}
     \\
     \rowfont{\scriptsize}Input &\rowfont{\scriptsize} EGNet &\rowfont{\scriptsize} MINet &\rowfont{\scriptsize} EGNet+cascadePSP &\rowfont{\scriptsize} MINet+cascadePSP &\rowfont{\scriptsize} MeticulousNet\_L+H
 \end{tabular}
 \captionof{figure}{Failure cases.}
 \label{fig: failure case}
\end{table*}

\section{Ablation Studies}
\label{sec: ablation}

\subsection{HierPR and Recursive Structure}

We first remove all our designed structures from MeticulousNet, and name it base network, which is a PSPNet~\cite{zhao2017pyramid} with ResNet50~\cite{he2016deep} backbone consisting of standard skip connections. Then, we add HierPR and recursive process separately into the base model. 
We evaluate their performances 
on the low resolution version of MOS600 
as shown 
in Table~\ref{Tab:ablation_component}. For HierPR, it increases both IoU, mBA and MQ by $0.17\%$, $1.45\%$ and $0.69\%$ respectively. The major improvements are on the boundary segmentation. For recursive process, it does not lead to a better IoU but improves mBA by $1.24\%$
leading to an increase of $0.58\%$ in MQ.
When integrating these two structures together, we observe a large improvements in the object body segmentation. The IoU is increased by $2.01\%$, which is not expected when analysing the structures separately. This indicates HierPR and recursive processes are good complementaries to be deployed together. Together, they also increase mBA and MQ by $1.93\%$ and $1.12\%$, respectively. In addition, we connect our decoder with MobileNetV3 based encoder~\cite{howard2019searching} leading to mobile\_MeticulousNet. 
The number of model parameters are decreased by more than $20$ times, but performance drop in MQ is only $0.73\%$. This proves the flexibility and adaptability of our decoder to be used with other encoders.

\renewcommand{\colwidthA}{3.5cm}
\renewcommand{\colwidthB}{0.95cm}
\renewcommand{\colwidthC}{1.1cm}
\begin{table}[t]
\centering{
\small
\setlength{\tabcolsep}{0.08cm}
\begin{tabular}{|C{\colwidthA}|C{\colwidthB}|C{\colwidthB}|C{\colwidthB}| C{\colwidthC} |}
\hline
Method & IoU $\uparrow$ & mBA $\uparrow$ & MQ $\uparrow$ & param.$\downarrow$ \\
\hline\hline
Base & $70.17$ & $58.79$ & $77.04$ & ---- \\
\hline
Base + HierPR  & $70.34$ & $60.24$ & $77.73$ & ---- \\
\hline
Base + Recur & $70.14$ & $60.03$ & $77.62$ & ---- \\
\hline
Base + HierPR + Recur & $\mathbf{72.18}$ & $\mathbf{60.72}$ & $\mathbf{78.16}$ & $70.30$ M \\
\hline
Mobile + HierPR + Recur & $71.42$ & $59.31$ & $77.43$ & $\;\mathbf{3.33}$ M \\
\hline
\end{tabular}}
\caption{Ablation studies on low resolution MOS600. Base and Mobile means PSPNet~\cite{zhao2017pyramid} encoder with ResNet50~\cite{he2016deep} and MobileNetV3~\cite{howard2019searching} backbones. (Units: $1\times 10^{-2}$)}
\label{Tab:ablation_component}
\end{table}

\renewcommand{\colwidthA}{3.0cm}
\renewcommand{\colwidthB}{1.0cm}
\renewcommand{\colwidthC}{1.1cm}
\begin{table}[t]
\centering{
\small
\setlength{\tabcolsep}{0.08cm}
\begin{tabular}{|C{\colwidthA}|C{\colwidthB}|C{\colwidthB}|C{\colwidthB}|}
\hline
Method & IoU $\uparrow$ & mBA $\uparrow$ & MQ $\uparrow$ \\
\hline\hline
HierPR\_$20\%$ & $71.23$ & $60.16$ & $77.79$ \\
\hline
HierPR\_$15\%$ & $71.34$ & $\mathbf{60.91}$ & $\mathbf{78.21}$ \\
\hline
HierPR\_$10\%$ & $\mathbf{72.18}$ & $60.72$ & $78.16$ \\
\hline
\end{tabular}}
\caption{Ablation studies on low resolution MOS600. We adopt HierPR\_10\% in our main experiments. (Units: $1\times 10^{-2}$)}
\label{Tab:ablation_settings}
\end{table}

\subsection{Other Settings}

In this part, we explore different settings for HierPR and a feasibility of the combination of an mobile encoder with the recursive decoder. In HierPR block, 
an area percentage $a\%$ is specified in which it is allowed to 
perform refinements. 
In our experiments, we set this value as $10\%$ and represent the block as HierPR\_$10\%$. 
We investigate whether larger area of local refinements will lead to better performances. Note that the extreme case $0\%$ means no HierPR is applied which is discussed previously. Evaluations are shown in Table~\ref{Tab:ablation_settings}. We do not observe a clear benefit brought by the increase of the refinement area. Comparing HierPR\_$15\%$ with HierPR\_$10\%$, although mBA is increased by $0.19\%$ but IoU is decreased by $0.84\%$, making the MQ comparable in these two cases. However, when the area percentage further grows up, it will harm the performance. 
HierPR\_$20\%$ has the worst performances. 

\section{Failure Case Analysis}

Our method is good at capturing small details of objects, but the drawback is paying too much attention on the color variations. 
This causes the failure cases of two categories: First, complicated textures or other high frequency noises make the predictions unstable, as shown in the first and second examples in Figure~\ref{fig: failure case}. Second, local color changes break the completeness of the overall object recognition. For example, in the last raw the wine glass in the hand prevent the model from recognizing the lady's shoulder and parts of arms. We propose MeticulousNet as the first baseline for Meticulous Object Segmentation and consider overcoming the excessive attention on color variations as one of the future development directions for MOS methods.

\section{Conclusion}
In this work, we propose a task named Meticulous Object Segmentation (MOS). MOS is dedicated to segment complex objects with finest details from high resolution images. This problem is essential in applied techniques such as image editing, but not well addressed in previous tasks or methods. By inviting the community to explore MOS methods, the goal of our work is to promote the emergence of novel algorithms with great applicable values.

To set up the benchmark method for MOS, we propose MeticulousNet containing a recursive decoder. The superior performances of our method compared with the combination of current state of the art models are experimentally demonstrated. We also invent Meticulosity Quality and release MOS600 dataset to evaluate MOS methods. 

{\small
\bibliographystyle{ieee_fullname}
\bibliography{egbib}
}

\end{document}